\documentclass{article}

\usepackage{pifont}

\newcommand{\ares}{\textsc{Ares}}

\usepackage{microtype}
\usepackage{graphicx}
\usepackage{subcaption}
\usepackage{booktabs} % for professional tables
\usepackage[utf8]{inputenc}
\usepackage[table]{xcolor} 
\usepackage{tabularx}
\usepackage{hyperref}
\usepackage{float}

\usepackage[preprint]{icml2026}

\usepackage{amsmath}
\usepackage{amssymb}
\usepackage{mathtools}
\usepackage{amsthm}
\usepackage{multirow}
\usepackage{makecell}

\usepackage{xcolor}
\usepackage[most]{tcolorbox}

\definecolor{GreenPigment}{rgb}{0.00,0.65,0.31}
\definecolor{DarkSalmon}{rgb}{0.91,0.59,0.48}
\definecolor{RedOrange}{rgb}{1,0.5,0}
\definecolor{BlueGreen}{rgb}{0.0, 0.5, 0.5}
\usepackage[svgnames]{xcolor}

\newcommand{\blue}[1]{${\color{BlueGreen}\downarrow #1}$}
\newcommand{\red}[1]{${\color{RedOrange}\uparrow #1}$}
\newcommand{\bluer}[1]{${\color{BlueGreen}\uparrow #1}$}
\newcommand{\redr}[1]{${\color{RedOrange}\downarrow #1}$}

\definecolor{bestblue}{RGB}{230, 240, 255} 
\definecolor{headergray}{RGB}{240, 240, 240} 
\definecolor{highlight}{RGB}{255, 245, 230}

\usepackage[capitalize,noabbrev]{cleveref}

\theoremstyle{plain}

\theoremstyle{definition}

\theoremstyle{remark}

\usepackage[textsize=tiny]{todonotes}
\usepackage{color}

\icmltitlerunning{{\ares}: Adaptive Reasoning Effort Selection for Efficient LLM Agents}

\begin{document}

\twocolumn[
  \icmltitle{{\ares}: Adaptive Reasoning Effort Selection for Efficient LLM Agents}

  % It is OKAY to include author information, even for blind submissions: the
  % style file will automatically remove it for you unless you've provided
  % the [accepted] option to the icml2026 package.

  % List of affiliations: The first argument should be a (short) identifier you
  % will use later to specify author affiliations Academic affiliations
  % should list Department, University, City, Region, Country Industry
  % affiliations should list Company, City, Region, Country

  % You can specify symbols, otherwise they are numbered in order. Ideally, you
  % should not use this facility. Affiliations will be numbered in order of
  % appearance and this is the preferred way.
  \icmlsetsymbol{equal}{*}

  \begin{icmlauthorlist}
    \icmlauthor{Jingbo Yang}{ucsb}
    \icmlauthor{Bairu Hou}{ucsb}
    \icmlauthor{Wei Wei}{accenture}
    \icmlauthor{Yujia Bao}{equal,accenture}
    \icmlauthor{Shiyu Chang}{equal,ucsb}
  \end{icmlauthorlist}

  \icmlaffiliation{ucsb}{Department of Computer Science, University of California, Santa Barbara}
  \icmlaffiliation{accenture}{Center for Advanced AI, Accenture}
  % \icmlaffiliation{apple}{Apple}

  \icmlcorrespondingauthor{Jingbo Yang}{jingbo@ucsb.edu}
  % \icmlcorrespondingauthor{Yujia Bao}{bao@yujia.io}
  \icmlcorrespondingauthor{Shiyu Chang}{chang87@ucsb.edu}
  % You may provide any keywords that you find helpful for describing your
  % paper; these are used to populate the "keywords" metadata in the PDF but
  % will not be shown in the document
  \icmlkeywords{Machine Learning, ICML}

  \vskip 0.3in
]

% this must go after the closing bracket ] following \twocolumn[ ...

% This command actually creates the footnote in the first column listing the
% affiliations and the copyright notice. The command takes one argument, which
% is text to display at the start of the footnote. The \icmlEqualContribution
% command is standard text for equal contribution. Remove it (just {}) if you
% do not need this facility.

% Use ONE of the following lines. DO NOT remove the command.
% If you have no special notice, KEEP empty braces:
\printAffiliationsAndNotice{\icmlEqualContribution}  % no special notice (required even if empty)
% Or, if applicable, use the standard equal contribution text:
% \printAffiliationsAndNotice{\icmlEqualContribution}

\begin{abstract}
Modern agents powered by thinking LLMs achieve high accuracy through long chain-of-thought reasoning but incur substantial inference costs. While many LLMs now support configurable reasoning levels (\emph{e.g.}, high/medium/low), static strategies are often ineffective: using low-effort modes at every step leads to significant performance degradation, while random selection fails to preserve accuracy or provide meaningful cost reduction. 
However, agents should reserve high reasoning effort for difficult steps like navigating complex website structures, while using lower-effort modes for simpler steps like opening a target URL.
In this paper, we propose {\ares}, a framework for per-step dynamic reasoning effort selection tailored for multi-step agent tasks. {\ares} employs a lightweight router to predict the lowest appropriate reasoning level for each step based on the interaction history. To train this router, we develop a data generation pipeline that identifies the minimum reasoning effort required for successful step completion. We then fine-tune the router to predict these levels, enabling plug-and-play integration for any LLM agents. We evaluate {\ares} on a diverse set of agent tasks, including \texttt{TAU-Bench} for tool use agents, \texttt{BrowseComp-Plus} for deep-research agents, and \texttt{WebArena} for web agents.
Experimental results show that {\ares} reduces reasoning token usage by up to 52.7\% compared to fixed high-effort reasoning, while introducing minimal degradation in task success rates.
\end{abstract}

\section{Introduction}
\label{sec:intro}

\begin{figure*}[t]
    \centering
    \includegraphics[width=\linewidth]{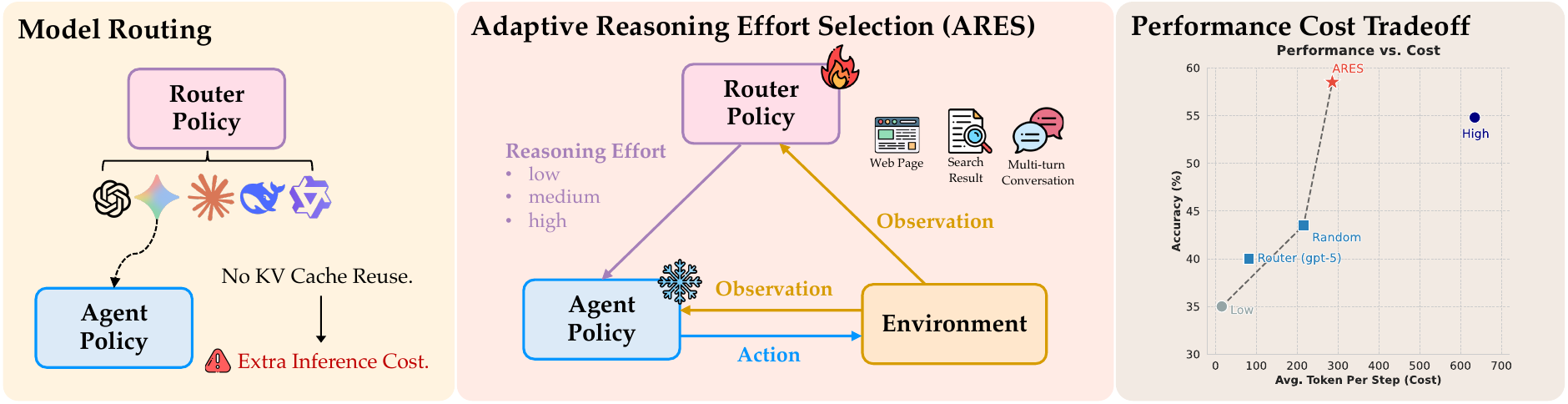} 
    
    \caption{
        \textbf{Overview of the Adaptive Reasoning Effort Selection (ARES) Framework.} 
        \textbf{Left:} Traditional Model Routing which often incurs extra inference costs without KV cache reuse. 
        \textbf{Middle:} Our proposed {\ares} framework, dynamically allocates reasoning effort at each step. 
        \textbf{Right:} {\ares} (red star) achieves the optimal balance between performance and cost compared to baselines.
    }
    \label{fig:intro}
\end{figure*}

Recent advancements in reasoning large language models (LLMs)~\citep{team2025kimi,guo2025deepseek,comanici2025gemini} have significantly boosted the capabilities of autonomous agents~\citep{jin2025search,sun2025scaling, liu2025learning,yang2025ultracua} in complex, multi-step decision-making tasks. By leveraging extended chain-of-thought (CoT) reasoning, these agents can perform deeper environment analysis and more rigorous planning before executing actions. However, this improved performance comes at a substantial inference cost, as a large number of reasoning tokens are accumulated at each step along the multi-step trajectories. To control these costs, a straightforward approach is to leverage the configurable ``thinking levels'' (or reasoning efforts) now supported by various state-of-the-art LLMs~\cite{singh2025openai}. These models (\emph{e.g.}, the \texttt{GPT-5} or \texttt{Gemini-3}) allow users to manually select from thinking modes, such as \textit{high/medium/low} or \textit{thinking/fast} modes, 
to balance performance and budget. With these options, users can configure LLM agents to always reason at lower levels at each step to reduce the cost.

However, such a uniform approach is often suboptimal because not all steps in a task require the same level of thinking. While some steps are simple, others demand intensive reasoning to avoid errors. Consequently, a fixed, static strategy often fails to balance performance and cost effectively. For example, a naive approach that consistently applies a low reasoning effort to minimize costs leads to severe performance degradation. As illustrated in Figure~\ref{fig:intro}, the agent powered by \texttt{gpt-oss-20b} suffers a nearly $20\%$ drop after switching the reasoning effort from ``high" to ``low" at every decision step. This performance gap suggests that a coarse-grained reduction in thinking effort is insufficient for maintaining agent effectiveness in complex environments.

To address these challenges, we propose a framework for dynamic reasoning effort allocation tailored for LLM agents. The core idea is to move beyond static configurations by adaptively determining the most suited reasoning effort for each individual step, reducing reasoning costs while preserving performance. While adaptive thinking has been explored in single-step tasks such as mathematical reasoning and competitive programming~\cite{shen2025dast,cui2025adaptive}, its application to LLM agents remains non-trivial. A slight reasoning deficit in an early step can lead to error propagation, making the balance between efficiency and long-term success far more complex. Furthermore, our approach distinguishes itself from existing model routing strategies~\cite{su2025toolorchestra,zhang2025router,amayuelas2025self}, where user inputs and tasks are routed to models with different sizes and capabilities. The trade-off between performance and cost is not always predictable or monotonic. In contrast, by leveraging the intra-model thinking levels (\textit{e.g.}, switching between a model's own high and low reasoning modes), we can ensure a more well-defined and consistent performance-cost frontier. This allows for a more granular control over the agent's behavior without the integration overhead of maintaining multiple heterogeneous models. In addition, this paradigm allows the agent to preserve and reuse the KV cache across different reasoning effort levels. This avoids the significant latency or computational costs associated with re-encoding context for a different model, thereby maximizing the token-saving benefits of our adaptive framework.

Under this setting, we propose {\ares}, a method that introduces a small-scale LM as a reasoning-effort router to work alongside the LLM agent. Specifically, the router takes the current interaction history as input and directly predicts the lowest appropriate reasoning-effort level for the next step. The agent then uses the predicted reasoning level to perform the next step. This design is model-agnostic and can be seamlessly integrated into any existing agent architecture.
During training, we develop a multi-phase automated data generation pipeline that identifies and labels the minimum reasoning effort required for each step within a trajectory. Specifically, we break down the multi-step reasoning effort labeling task as a combination of single-turn classification problems. We start with collecting high-quality trajectory data to obtain the ground truth agent actions. After that, we annotate the lowest reasoning effort which produces the correct action stably. Conditioned on the effort prediction, we induce a rationale for analyzing the environment and predicting the effort. 
We then fine-tune a lightweight router (e.g., \texttt{Qwen3-1.7B}) to predict these effort levels by minimizing the next-token prediction loss.

We validate {\ares} through comprehensive experiments across diverse agent tasks, including tool-use agents, deep-research agents, and web agents. Specifically, we fine-tune a lightweight \texttt{Qwen3-1.7B} model as the reasoning-effort router in {\ares}. Compared to consistently using high reasoning effort, {\ares} substantially reduces token usage while maintaining task performance. For example, on \texttt{TAU-Bench}, {\ares} achieves an 52.7\% reduction in reasoning cost and even slightly improves performance relative to always using high reasoning effort.

\section{Related Work}
In this section, we provide an overview of prior works which closely relate to or partially motivated our work, including LLM routing, efficient reasoning.

\paragraph{LLM Routing.}
Routing across models of varying scales or architectures is a well-established technique for balancing performance and cost. Early methods~\citep{srivatsa2024harnessing,ong2024routellm,feng2024graphrouter} trained routers—often using human preference data~\citep{chiang2024chatbot}—to select the optimal model per task. Other approaches employ clustering-based routing, such as Avengers~\citep{zhang2025avengers}, BESTRoute~\citep{ding2025best}, and related frameworks~\citep{jitkrittum2025universal}. However, these primarily target single-turn tasks (\emph{e.g.}, math), reducing routing to an independent classification problem. In contrast, \textsc{Ares} addresses the more complex multi-turn decision-making setting where steps are inherently interdependent.

Recent research has extended model routing to multi-turn agent tasks. ToolOrchestra~\citep{su2025toolorchestra} utilizes an orchestrator LLM for joint planning and routing, though it relies on subjective natural language descriptions of model capabilities. Router-R1~\citep{zhang2025router} frames multi-turn routing as a sequential process but focuses on simple QA tasks (\emph{e.g.}, NQ~\citep{kwiatkowski2019natural}, TriviaQA~\citep{joshi2017triviaqa}) that often collapse into single-turn RAG problems. More recently, EvoRoute~\citep{zhang2026evoroute} proposed an experience-driven self-routing framework. While these multi-model paradigms show promise, they remain constrained by non-monotonic cost-performance relationships and inefficiencies from redundant context encoding, as detailed in Section~\ref{sec:intro}. In contrast, our setting is a better defined optimization problem, and reuses the KV cache to avoid additional inference cost.

\paragraph{Efficient and adaptive LLM reasoning.} 
Our method is also related to recent approaches for adaptive thinking of LLMs, which aim to teach models to reason adaptively according to the input difficulty~\citep{liu2025diffadapt,yu2025think,wu2025efficiency,zhang2025dart} or a user-specified thinking budget~\citep{huang2025adactrl,alomrani2025reasoning,hou2025thinkprune}. 
Similar to other works on model routing, these approaches primarily focus on controlling reasoning in a single-turn setting, by dynamically adjusting the length of reasoning traces or truncating intermediate thoughts under a given budget. 
In contrast, our approach models reasoning effort as a sequential decision-making process, making it applicable to more complex, multi-turn agent tasks.

\section{Method}
In this section, we present {\ares}, a framework that dynamically allocates most efficient but effective reasoning effort across decision steps. We first provide the problem formulation of optimizing router model, and then elaborate how we train an effective reasoning effort router using both supervised fine-tuning (SFT) as well as reinforcement learning (RL). 

\subsection{Problem Formulation}
\label{sec: formulation}
To formalize the decision-making process of the router, we first define the agent's task and interaction environment. An agent task is characterized by a goal or user query $x \in \mathcal{X}$. The interaction proceeds over discrete turns $t = 1, \dots, T$. At each turn $t$, the environment provides an observation $o_t \in \mathcal{O}$ (\emph{e.g.}, external tool outputs or web page content). The LLM agent $\mathcal{M}_{\text{agent}}$ with parameter $\phi$ first performs chain-of-thought reasoning and then predicts the next action $a_t$. 
Notably, the reasoning level $e_t$ of the LLM is configurable and can be selected from a fixed set, such as high/medium/low supported by \texttt{gpt-oss-20b}.
Let $h_t = (x, o_1, a_1, \dots, o_{t-1}, a_{t-1})$ denote the interaction history up to turn $t$, the agent model $\mathcal{M}_{\text{agent}}$ with parameters $\phi$ predicts the next action $a_t$:
\begin{equation}
    a_t \sim P_\phi(a_t \mid h_t, o_t, e_t)
\end{equation}
The agent's objective is to produce a sequence of actions such that the complete trajectory $\tau = (x, o_1, a_1, \dots, o_T, a_T)$ satisfies the task success criterion, denoted by a verification function $\mathcal{V}(\tau) \in \{0, 1\}$.

The router is a lightweight LLM $\mathcal{M}_{\text{router}}$ with parameters $\theta$. At each turn $t$, the router receives the same input context as the agent—the task, history, and current observation—and predicts the optimal reasoning effort level $e_t$ from a discrete space $\mathcal{E} = \{ e_{\text{low}}, e_{\text{mid}}, e_{\text{high}} \}$:
$e_t = \mathcal{M}_{\text{router}}(h_t, o_t; \theta)$.
The selected $e_t$ represents the reasoning strategy that $\mathcal{M}_{\text{agent}}$ will employ for that specific turn.

The objective of the router is to minimize the cumulative inference cost across the trajectory while ensuring that the task remains successful.
We quantify the cost of each step, $\mathrm{cost}(e_t)$, as the total number of tokens generated by the agent at turn $t$, which includes both the internal reasoning (thinking) tokens and the tokens for the final action. 
Formally, the optimization goal is to find a selection policy that maximizes the task success rate while minimizing the cumulative computational cost:
\begin{equation}
    \max_{\theta} \mathbb{E}_{x \sim \mathcal{X}, \tau \sim \mathcal{T}(\theta, \phi)} \left[ \mathcal{V}(\tau, x) - \lambda \sum_{t=1}^{T} \text{cost}(e_t) \right]
\end{equation}
where $\tau \sim \mathcal{T}(\theta, \phi)$ is the trajectory distribution on task $x$ induced by the LLM agent parametrized by $\phi$ and the router parameterized by $\theta$. $\mathcal{V}(\tau, x)$ measures whether the task success of trajectory $\tau$ on task $x$. By solving this optimization problem, the router learns to dynamically allocate minimal reasoning resources at each step while preserving the final task utility.

\begin{figure*}[t]
    \centering
    \includegraphics[width=\linewidth]{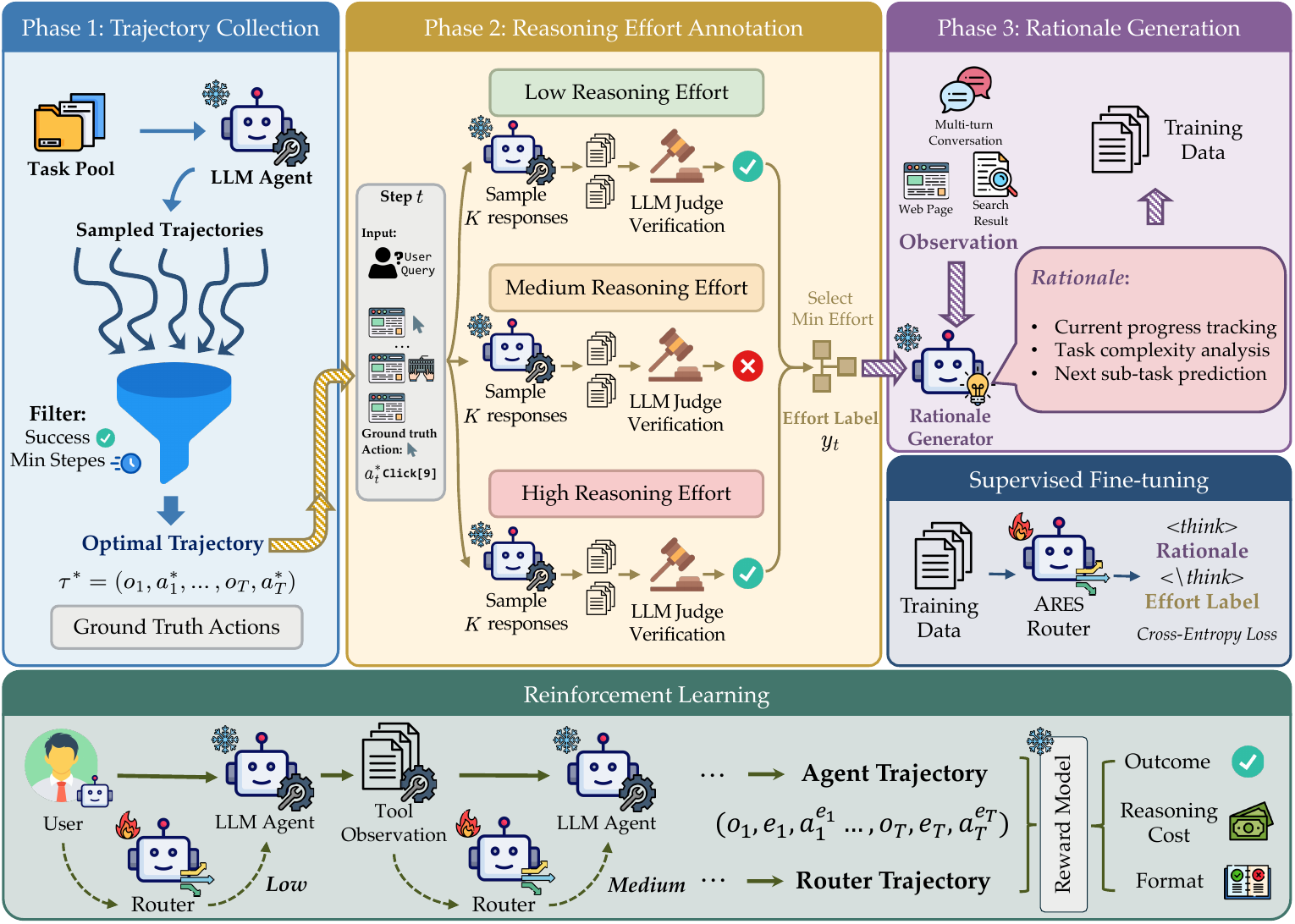} 
    
    \caption{\textbf{Overview of the {\ares} training pipeline.} 
(1) Trajectory Collection: Optimal ground-truth paths are defined by filtering successful trajectories with minimal steps. 
(2) Effort Annotation: The minimum sufficient reasoning effort for each step is identified via sampling and LLM verification. 
(3) Rationale Generation: A teacher LLM generates semantic justifications based on task observations and complexity. 
(4) Supervised Fine-tuning: The {\ares} router is fine-tuned to jointly predict rationales and effort labels.
(5) Reinforcement Learning: The fine-tuned {\ares} router will be further trained using GRPO with outcome, reasoning cost, and format reward.}
    \label{fig:method}
\end{figure*}

\subsection{Supervised Fine-tuning Pipeline}
\label{sec:method}
The primary challenge in training a reasoning effort router lies in label acquisition. Unlike standard classification tasks, the optimal reasoning effort for a specific step cannot be directly observed. Obtaining this label through naive trial-and-error is computationally prohibitive due to the $|\mathcal{E}|^T$ search space. Moreover, a suboptimal effort allocation in early steps can lead to error propagation, making it difficult to directly optimize for efficiency across a trajectory.

To resolve this, our core idea is to decouple the reasoning effort selection from the task-solving trajectory. Instead of searching for everything at once, we first find a successful trajectory and then test each step one by one to find the minimum thinking effort needed to get that specific step right. This allow us to identify the lowest sufficient reasoning effort required to reproduce a correct action without the noise of error propagation.

\paragraph{Phase 1: Trajectory Collection.}

Rather than searching for efficiency across arbitrary action sequences, we anchor our analysis on high-quality, successful trajectories. For each training task $x$, we sample $N$ successful trajectories using the agent $\mathcal{M}_{\text{agent}}$ under the maximum effort level $e_{\text{high}}$. From these successful trials, we select the most concise trajectory $\tau^* = (o_1, a^*_1, \dots, o_T, a^*_T)$ to serve as the reference path. The rationale for this selection is two-fold. First, trajectories with more steps inflate the total reasoning cost and make it harder to identify the true minimum effort required for the core task. In contrast, concise trajectories provide a cleaner signal of the essential reasoning needed. Second, by fixing a high-quality action sequence, we transform the long-horizon optimization problem into a series of independent step-wise labeling tasks. This allows us to precisely measure how different reasoning levels affect the execution of each ground-truth action $a^*_t$ in isolation.

We also note that the {\ares} framework is also compatible with external data sources. The router can be trained using existing open-source agent datasets, even if those trajectories were generated by a different LLM. This flexibility allows {\ares} to leverage high-quality interaction data without always requiring expensive self-sampling. As demonstrated in our experiments (Section~\ref{sec:exp}), the router generalizes effectively across different interaction styles and model sources.

\paragraph{Phase 2: Reasoning Effort Annotation.}

With the optimal trajectory $\tau^* = (o_1, a^*_1, \dots, o_T, a^*_T)$ as a reference, we decompose it into individual decision steps to identify the minimum required reasoning effort for each action. For each step $t$, we treat the action $a^*_t$ as the ground truth. Our goal is to find the lowest effort level $e \in \mathcal{E}$ that can reliably reproduce $a^*_t$ given the history $h_t$ and observation $o_t$.

To account for the inherent stochasticity of LLMs, we employ a multi-trial verification process. For each effort level $e \in \{e_{\mathrm{low}}, e_{\mathrm{mid}}, e_{\mathrm{high}}\}$, we sample the agent's response $K$ times ($K=3$ in our experiments) and compare the predicted actions $\{\hat{a}_{t,k}^{(e)}\}_{k=1}^K$ against the ground truth $a^*_t$.
An effort level $e$ is considered \textit{sufficient} for step $t$ if it reliably reproduces the correct action in a majority of trials. Specifically, we define a verification function $\mathcal{V}(\hat{a}, a^*_t)$ that returns 1 if the predicted action is functionally equivalent to the ground truth and 0 otherwise. We require the action to be correct in at least $M$ out of $K$ trials. The candidate set of sufficient levels $\mathcal{C}_t$ is then defined as:
\begin{equation}
\mathcal{C}_t = \left\{ e \in \mathcal{E} \mid \sum_{k=1}^{K} \mathcal{V}(\hat{a}_{t,k}^{(e)}, a^*_t) \ge M \right\}
\end{equation}
To ensure the verification process $\mathcal{V}$ is both rigorous and practical, we tailor the evaluation criteria to diverse agent domains. Specifically, two actions are considered functionally equivalent if they produce the same effect on the environment. 
For tool-use agents, an action is valid if the predicted tool name matches $a^*_t$ and the key parameters are identical. 
For web agents, an action is considered correct if it interacts with the environment in exactly the same manner as $a^*_t$ (\textit{e.g.}, clicking a specific element such as \texttt{click[1316]}). 
For deep research agents, where the primary action involves searching, we utilize an LLM judge to determine if two search queries are semantically equivalent. 
Across all three agent types, when the agent outputs natural language messages (\textit{e.g.}, soliciting user information or providing a final answer), we also employ an LLM judge to verify that the messages are semantically consistent.

Finally, we assign the label $y_t$ by picking the lowest-cost sufficient level. For example, if both $e_{\mathrm{low}}$ and $e_{\mathrm{mid}}$ are sufficient, we label the step as $e_{\mathrm{low}}$ to maximize efficiency. If no level passes the robustness check, the step is discarded to maintain data quality. This exhaustive labeling process creates a high-quality supervision signal that defines the computational lower bound for the task.

\paragraph{Phase 3: Rationale Generation.}

To enhance the router's predictive accuracy, we train it to generate a brief reasoning rationale before outputting the final effort label. This approach allows the lightweight router to analyze the current state explicitly rather than performing a direct mapping from context to label. We compare this design against a baseline that performs direct label classification in Section~\ref{subsec:ablation} to evaluate the empirical benefits of rationale generation.

To synthesize high-quality training data for this process, we employ a powerful teacher model to perform post-hoc analysis on the labeled trajectories. For each step $t$, the teacher model is provided with the interaction history, the current observation, and the ground-truth label $y_t$ derived from Stage 2. It is then tasked with justifying why $y_t$ is the most appropriate level for the upcoming action. The generated rationale integrates several key factors: it assesses the complexity of the latest observation, evaluates the current progress of the overall task, and estimates the inherent difficulty of the required sub-task. To ensure that the router's own reasoning does not introduce significant latency, we impose a strict length constraint on the teacher model, limiting the rationale to a concise summary of 3--5 sentences. 
This ensures that the router remains lightweight and fast, providing a favorable trade-off between its own token overhead and the substantial token savings it enables for the agent.

\paragraph{Supervised Fine-tuning.}

Finally, we fine-tune a lightweight model to serve as the reasoning router using the augmented dataset $\mathcal{D} = \{ (h_t, o_t, r_t, y_t) \}$. The model is trained with a standard next-token prediction objective, where it learns to generate the reasoning rationale $r_t$ followed by the discrete effort label $y_t$ given the interaction context.

\subsection{Reinforcement Learning}

While SFT trains the \textsc{Ares} router to maximize the probability of selecting the minimum sufficient reasoning effort at each step, the SFT data synthesis pipeline reduces the multi-turn routing process to a series of single-step decisions with a greedy objective. This approach may fail to capture the complex multi-turn dynamics of the routing process. Specifically, the greedy nature of SFT introduces two primary shortcomings: \ding{182} single-step selection assumes that all previous effort levels were optimal, leaving the router without training signals on how to recover from sub-optimal selections made in prior turns; \ding{183} the current SFT data provides only one step-wise optimal selection per query, lacking the contrastive signals necessary to optimize the total query-wise reasoning effort across different selection sequences. To address these challenges and fully realize the potential of the \textsc{Ares} router, we further explore training via reinforcement learning (RL).

\paragraph{Router Rollout.} 
Following the formulation in Section~\ref{sec: formulation}, at each agent decision turn $t$, the \textsc{Ares} router takes the trajectory history $h_t$ and the current observation $o_t$ as input to produce: (1) a rationale $r_t$ (chain-of-thought), where the router analyzes the current task progress and the difficulty of the next sub-task; and (2) a reasoning effort prediction $e_t$. This selected effort level is then provided as part of the input to the downstream agent to generate the subsequent action $a^{e_t}_t$, as is shown in Figure~\ref{fig:method}. The rollout continues until the task is successfully completed or the agent reaches the maximum number of allowed steps.

\paragraph{Reward Design.} 
To guide the router toward balancing task success with computational efficiency, we define a composite reward function $R(\tau)$ for a trajectory $\tau$ of length $T$, and employ Group Relative Policy Optimization (GRPO)~\citep{shao2024deepseekmath} for training. The total reward is the sum of three components:

\begin{equation}
    R(\tau) = 
    \begin{cases}
    R_{\text{out}} + R_{\text{cost}} + R_{\text{form}} & \text{if task is successful} \\
    R_{\text{out}} + R_{\text{form}} & \text{otherwise}
    \end{cases}   
\end{equation}

\ding{182} \textbf{Outcome Reward ($R_{\text{out}}$):} Upon completion of the rollout, we verify the final environment state to determine task success. To ensure that the router prioritizes task completion, we assign a high-magnitude reward:
\begin{equation}
    R_{\text{out}} = 
    \begin{cases} 
    +5.0 & \text{if task is successful} \\
    0.0 & \text{otherwise}
    \end{cases}
\end{equation}

\ding{183} \textbf{Reasoning Cost Reward ($R_{\text{cost}}$):} At each turn $t$, choosing a specific effort level $e_t$ incurs a penalty $c(e_t)$ to discourage redundant computation. We define the turn-wise cost as:
\begin{equation}
    c(e_t) = 
    \begin{cases} 
    -0.2 & \text{if } e_t = e_{\text{low}} \\
    -0.5 & \text{if } e_t = e_{\text{mid}} \\
    -1.0 & \text{if } e_t = e_{\text{high}}
    \end{cases}
\end{equation}
The total cost reward is normalized as the trajectory-average: $R_{\text{cost}} = \frac{1}{T} \sum_{t=1}^{T} c(e_t)$. The reasoning cost penalty is exclusively applied to successful trajectories. This prevents the router from learning a degenerate policy where it deliberately selects low-effort actions to fail difficult tasks quickly, merely to minimize accumulated step-wise penalties.

\ding{184} \textbf{Format Reward ($R_{\text{form}}$):} To ensure the router follows the prescribed reasoning template (e.g., encapsulating rationales within \texttt{<think>} tags), we apply a penalty of $R_{\text{form}} = -1.0$ if the output format is violated. In such cases, the rollout is immediately terminated and marked as a failure, as we empirically found routing a format mistake to a fixed reasoning effort will make the training unstable.

This structured reward encourages the router to first prioritize the dominant outcome signal ($R_{\text{out}}$) to ensure correctness, before optimizing the efficiency-accuracy trade-off via the cost penalty.

\paragraph{RL Data Filtering.}
To ensure the \textsc{Ares} router learns an effective reasoning effort selection strategy, the quality of the training signals is paramount. We implement a targeted data processing pipeline to filter the initial prompt pool and highlight the most informative prompts for optimization.

For each prompt in our training set, we perform $N$ rollouts (\emph{e.g.}, $N$=8) using the SFT-initialized router to observe the variance in agent performance and computational cost. We then apply the following two-stage filtering process:

\ding{182} \textbf{Zero-Success Filtering:} We calculate the success rate (SR) for each prompt across all $N$ rollouts. We discard any prompt where SR=0, as these tasks likely exceed the fundamental capabilities of the underlying agent LLM regardless of the reasoning effort applied. Including these ``unsolvable" samples would introduce significant noise into the training signal, as the router might incorrectly attribute the inherent failures of the agent to its own effort selection.

\ding{183} \textbf{Variance-Based Efficiency Selection:} For the remaining prompts where the agent achieves a perfect success rate (SR=100\%), we calculate the total reward for each rollout and determine the variance across the N samples of each prompt. We retain only those prompts whose reward variance falls within the top 30\%. The rationale is that high-variance rewards on successful prompts indicate that multiple reasoning strategies achieve the same outcome but with vastly different costs. These samples provide the strongest signal for the router to learn how to minimize reasoning tokens without sacrificing task accuracy.

By curating the data in this manner, we ensure the reinforcement learning process focuses on prompts where the reasoning effort selection is the primary driver of the efficiency-accuracy trade-off.

\section{Experiments}
\label{sec:exp}
\subsection{Experimental Setup}

\begin{table*}[t]
  \centering
  \small
  \setlength{\tabcolsep}{4.0pt} 
  \renewcommand{\arraystretch}{1.0}
  
  \caption{Main evaluation results in \textbf{TAU-Bench}, \textbf{BrowserComp-Plus}, and \textbf{WebArena}, using \texttt{gpt-oss-20b} as the backbone LLM. Arrows ($\uparrow/\downarrow$) indicating an increase or decrease in absolute value.
  \textcolor{RedOrange}{Orange} and \textcolor{BlueGreen}{blue} indicate beneficial and detrimental changes.} 

  \resizebox{0.95\linewidth}{!}{
  \begin{tabular}{cl|cc|c|cc|cc|cc}
    \toprule
    \rowcolor{CadetBlue!15}
    \cellcolor{white} & \cellcolor{white} 
    & \multicolumn{3}{c|}{\textbf{Performance}} 
    & \multicolumn{6}{c}{\textbf{Efficiency (Token Consumption)}} \\ 
    \cmidrule{3-5} \cmidrule{6-11}
    
    % \multirow{-2.25}{*}{\shortstack{\textbf{Method} \\ \textbf{Category}}} 
    % & \multirow{-2.25}{*}{\shortstack{\textbf{Routing} \\ \textbf{Method}}} 
    & \multirow{-2.25}{*}{\shortstack{\textbf{Method}}} 
    & \textbf{Acc. (\%)} & \hspace{5pt} $\Delta_{\mathrm{Acc}}$ \hspace{5pt} & \hspace{15pt}\textbf{$S$} \hspace{15pt}
    & \hspace{10pt}\textbf{$T_{\text{total}}$} \hspace{10pt}& $\Delta_{\mathrm{token}}$ 
    & \hspace{10pt}\textbf{$T_{\text{task}}$} \hspace{10pt} & $\Delta_{\mathrm{token}}$ 
    & \hspace{10pt}\textbf{$T_{\text{step}}$} \hspace{10pt}& $\Delta_{\mathrm{token}}$ \\
    \midrule
    \multicolumn{11}{c}{\textbf{TAU-Bench}  \textit{Retail}} \\
    \midrule
    % \multirow{4}{*}{\textsc{Rule-based}} 
    \multirow{4}{*}{\makecell{\textsc{Rule}\\\textsc{-based}}} 
    & Low & 35.0 & \red{19.8} & 13.5 & 25k & \bluer{627k} & 223 & \bluer{5454} & 15 & \bluer{373}\\
    & Medium & 47.3 & \red{7.5} & 14.5 & 137k & \bluer{515k} & 1198 & \bluer{4479} & 82 & \bluer{306}\\
    & High & 54.8 & \blue{0.0} & 13.8 & 1007k & \redr{355k} & 8756 & \redr{3079} & 634 & \redr{246} \\
    & Random & 43.5 & \red{11.3} & 14.5 & 359k & \bluer{293k} & 3126 & \bluer{2550} & 215 & \bluer{172} \\
    \midrule
    \multirow{2}{*}{\makecell[c]{\textsc{Prompting}\\\textsc{-based}}}
    & GPT 5 & 40.0 & \red{14.8} & 14.0 & 132k & \bluer{520k} & 1148 & \bluer{4529} & 81 & \bluer{307} \\
    & Gemini 3 Pro & 46.1 & \red{8.7} & 14.7 & 128k & \bluer{524k} & 1115 & \bluer{4562} & 76 & \bluer{312} \\
    \midrule
    \rowcolor{gray!10}
    % \textsc{Ours} 
    & \textsc{Ares} & 54.8 & -- & 14.6 & 652k & -- & 5677 & -- & 388 & --\\
    \midrule
    \multicolumn{11}{c}{\textbf{TAU-Bench}  \textit{Airline}} \\
    \midrule
    % \multirow{4}{*}{\textsc{Rule-based}} 
    \multirow{4}{*}{\makecell{\textsc{Rule}\\\textsc{-based}}} 
    & Low & 32.0 & \red{4.0} & 10.8 & 12k & \bluer{666k} & 250 & \bluer{13327} & 23 & \bluer{1077} \\
    & Medium & 42.0 & \blue{6.0} & 13.2 & 98k & \bluer{580k} & 1961 & \bluer{11616} & 148 & \bluer{952}\\
    & High & 38.0 & \blue{2.0} & 10.6 & 873k & \redr{195k} & 17472 & \redr{3895} & 1654 & \redr{554} \\
    & Random & 34.0 & \red{2.0} & 13.5 & 521k & \bluer{157k} & 10437 & \bluer{3140} & 770 & \bluer{330} \\
    \midrule
    % \multirow{2}{*}{\shortstack[l]{\textsc{Prompting-based}\\\textsc{Router}}} 
    \multirow{2}{*}{\makecell[c]{\textsc{Prompting}\\\textsc{-based}}}
    & GPT 5 & 34.0 & \red{2.0} & 12.4 & 396k & \bluer{282k} & 7927 & \bluer{5650} & 641 & \bluer{459} \\
    & Gemini 3 Pro & 36.0 & \red{0.0} & 12.6 & 239k & \bluer{439k} & 4779 & \bluer{8798} & 379 & \bluer{721} \\
    \midrule
    \rowcolor{gray!10}
    % \textsc{Ours} 
    & \textsc{Ares} & 36.0 & -- & 12.3 & 678k & -- & 13577 & -- & 1100 & -- \\
    \midrule
    \multicolumn{11}{c}{\textbf{BrowseComp-Plus}} \\
    \midrule
    % \multirow{4}{*}{\textsc{Rule-based}} 
    \multirow{4}{*}{\makecell{\textsc{Rule}\\\textsc{-based}}} 
    & Low & 8.00 & \red{33.3} & 4.6 & 5k & \bluer{1066k} & 332 & \bluer{6449} & 72 & \bluer{113} \\
    & Medium & 34.0 & \red{7.3} & 26.2 & 538k & \bluer{533k} & 3590 & \bluer{3191} & 137 & \bluer{48} \\
    & High & 42.7 & \blue{1.4} & 55.4 & 1841k & \redr{770k} & 12276 & \redr{5495} & 220 & \redr{35} \\
    & Random & 30.7 & \red{10.6} & 18.0 & 392k & \bluer{679k} & 2616 & \bluer{4165} & 145 & \bluer{40} \\
    \midrule
    % \multirow{2}{*}{\shortstack[l]{\textsc{Prompting-based}\\\textsc{Router}}} 
    \multirow{2}{*}{\makecell[c]{\textsc{Prompting}\\\textsc{-based}}}
    & GPT 5 & 38.7 & \red{2.6} & 45.1 & 1398k & \redr{327k} & 9321 & \redr{2540} & 206 & \redr{21} \\
    & Gemini 3 Pro & 37.3 & \red{4.0} & 41.6 & 1144k & \redr{73k} & 7628 & \redr{847} & 183 & \bluer{2} \\
    \midrule
    \rowcolor{gray!10}
    % \textsc{Ours} 
    & \textsc{Ares} & 41.3 & -- & 36.5 & 1071k & -- & 6781 & -- & 185 & -- \\ 
    \midrule
    \multicolumn{11}{c}{\textbf{WebArena}} \\
    \midrule
    % \multirow{4}{*}{\textsc{Rule-based}} 
    \multirow{4}{*}{\makecell{\textsc{Rule}\\\textsc{-based}}} 
    & Low & 37.4 & \red{9.1} & 8.9 & 67k & \bluer{1445k} & 520 & \bluer{11203} & 58 & \bluer{1266} \\
    & Medium & 42.6 & \red{3.9} & 9.9 & 538k & \bluer{974k} & 4170 & \bluer{7553} & 420 & \bluer{904} \\
    & High & 45.0 & \red{1.5} & 10.0 & 2763k & \redr{1251k} & 21424 & \redr{9701} & 2154 & \redr{830} \\
    % \midrule
    & Random & 40.3 & \red{6.2} & 8.0 & 857k & \bluer{655} & 6643 & \bluer{5080} & 830 & \bluer{494} \\
    \midrule
    % \multirow{2}{*}{\shortstack[l]{\textsc{Prompting-based}\\\textsc{Router}}} 
    \multirow{2}{*}{\makecell[c]{\textsc{Prompting}\\\textsc{-based}}}
    & GPT 5 & 41.1 & \red{5.4} & 9.0 & 1159k & \bluer{353k} & 8990 & \bluer{2733} & 1001 & \bluer{323} \\
    & Gemini 3 Pro & 41.9 & \red{4.6} & 9.1 & 1164k & \bluer{348k} & 9023 & \bluer{2670} & 995 & \bluer{329} \\
    \midrule
    \rowcolor{gray!10}
    % \textsc{Ours} 
    & \textsc{Ares} & 46.5 & -- & 8.9 & 1512k & -- & 11723 & -- & 1324 & -- \\ 
    \bottomrule
  \end{tabular}
  }
  
  \label{tab:main}
\end{table*}

\paragraph{Environments.}
We evaluate {\ares} across three diverse agent environments to demonstrate its effectiveness and generalizability in real-world settings: \ding{182} \texttt{TAU-Bench}~\citep{yao2024tau} for tool-use agents, \ding{183} \texttt{BrowseComp-Plus}~\citep{chen2025browsecomp} for deep-research agents, and \ding{184} \texttt{WebArena}~\citep{zhou2023webarena} for web agents.

\texttt{TAU-Bench} evaluates agent tool-use within task-oriented dialogues across \emph{retail} and \emph{airline} domains. Each sample involves an interaction between a user simulator and an LLM agent, where the agent must execute database tool calls to resolve queries or complete tasks (\emph{e.g.}, flight bookings). Success is measured by the correctness of the final database state or the provided information. We apply \textsc{Ares} after every user message and tool observation.

\texttt{BrowseComp-Plus}, a controlled derivative of \texttt{BrowseComp}~\citep{wei2025browsecomp}, ensures experiment reproducibility by replacing open-web search with a fixed corpus and incorporating human verification. The benchmark evaluates agents in a multi-step setting involving interleaved reasoning and retrieval. At each step, the agent generates a reasoning sequence followed by a search action or final answer. \textsc{Ares} is invoked prior to each reasoning and action step.

\texttt{WebArena} provides functional websites across domains such as e-commerce, social forums, and software development. Agents receive webpage observations (\emph{e.g.}, accessibility trees) and perform human-like actions (\emph{e.g.}, click, scroll, type) to fulfill queries requiring information seeking, navigation, or configuration. \textsc{Ares} is integrated before every agent inference step.

\paragraph{Implementation.}
In our experiments, we use the \texttt{gpt-oss-20b} model as the backbone LLM for agents in each task and employ \texttt{GPT-4o} as the user simulator model in \texttt{TAU-Bench}, following the original implementation. For the deep-research agent, we use \texttt{BM25} as the retriever and utilize the original \texttt{gpt-oss} web search and browsing tools. For web agents, we adopt \textsc{AgentOccam} as the baseline agent implementation.~\citep{yang2024agentoccam}

\begin{table*}[t]
  \centering
  \small
  \setlength{\tabcolsep}{4.0pt} 
  \renewcommand{\arraystretch}{1.0}
  
  \caption{RL results in \textbf{TAU-Bench} using \texttt{gpt-oss-20b} as the backbone LLM. Arrows ($\uparrow/\downarrow$) indicating an increase or decrease in absolute value.
  \textcolor{RedOrange}{Orange} and \textcolor{BlueGreen}{blue} indicate beneficial and detrimental changes.} 

  \resizebox{0.95\linewidth}{!}{
  \begin{tabular}{cl|cc|c|cc|cc|cc}
    \toprule
    \rowcolor{CadetBlue!15}
    \cellcolor{white} & \cellcolor{white} 
    & \multicolumn{3}{c|}{\textbf{Performance}} 
    & \multicolumn{6}{c}{\textbf{Efficiency (Token Consumption)}} \\ 
    \cmidrule{3-5} \cmidrule{6-11}
    
    & \multirow{-2.25}{*}{\shortstack{\textbf{Method}}} 
    & \textbf{Acc. (\%)} & \hspace{5pt} $\Delta_{\mathrm{Acc}}$ \hspace{5pt} & \hspace{15pt}\textbf{$S$} \hspace{15pt}
    & \hspace{10pt}\textbf{$T_{\text{total}}$} \hspace{10pt}& $\Delta_{\mathrm{token}}$ 
    & \hspace{10pt}\textbf{$T_{\text{task}}$} \hspace{10pt} & $\Delta_{\mathrm{token}}$ 
    & \hspace{10pt}\textbf{$T_{\text{step}}$} \hspace{10pt}& $\Delta_{\mathrm{token}}$ \\
    \midrule
    \multicolumn{11}{c}{\textbf{TAU-Bench}  \textit{Retail}} \\
    \midrule
    % \multirow{4}{*}{\textsc{Rule-based}} 
    \multirow{4}{*}{\makecell{\textsc{Rule}\\\textsc{-based}}} 
    & Low & 35.0 & \red{23.5} & 13.5 & 25k & \bluer{451k} & 223 & \bluer{3918} & 15 & \bluer{271}\\
    & Medium & 47.3 & \red{11.2} & 14.5 & 137k & \bluer{339k} & 1198 & \bluer{2943} & 82 & \bluer{204}\\
    & High & 54.8 & \red{3.7} & 13.8 & 1007k & \redr{531k} & 8756 & \redr{4615} & 634 & \redr{348} \\
    & Random & 43.5 & \red{15.0} & 14.5 & 359k & \bluer{117k} & 3126 & \bluer{1015} & 215 & \bluer{71} \\
    \midrule
    % \multirow{2}{*}{\makecell[c]{\textsc{Prompting}\\\textsc{-based}}}
    % & GPT 5 & 40.0 & \red{14.8} & 14.0 & 132k & \bluer{520k} & 1148 & \bluer{4529} & 81 & \bluer{307} \\
    % & Gemini 3 Pro & 46.1 & \red{8.7} & 14.7 & 128k & \bluer{524k} & 1115 & \bluer{4562} & 76 & \bluer{312} \\
    % \midrule
    % \rowcolor{gray!10}
    % \multirow{2}{*}{{\ares}} 
    & \textsc{SFT} & 54.8 & \red{3.7} & 14.6 & 652k & \redr{176k} & 5677 & \redr{1536} & 388 & \redr{102}\\
    \rowcolor{gray!10}
    \multirow{-2}{*}{{\ares}} 
    & \textbf{\textsc{RL}} & 58.5 & -- & 14.4 & 476k & -- & 4141 & -- & 286 & -- \\
    \midrule
    \multicolumn{11}{c}{\textbf{TAU-Bench}  \textit{Airline}} \\
    \midrule
    \multirow{4}{*}{\makecell{\textsc{Rule}\\\textsc{-based}}} 
    & Low & 32.0 & \red{10.0} & 10.8 & 12k & \bluer{121k} & 250 & \bluer{2403} & 23 & \bluer{208} \\
    & Medium & 42.0 & \blue{0.0} & 13.2 & 98k & \bluer{35k} & 1961 & \bluer{692} & 148 & \bluer{83}\\
    & High & 38.0 & \red{4.0} & 10.6 & 873k & \redr{740k} & 17472 & \redr{14819} & 1654 & \redr{1423} \\
    & Random & 34.0 & \red{8.0} & 13.5 & 521k & \redr{388k} & 10437 & \redr{7784} & 770 & \redr{539} \\
    % \midrule
    % \multirow{2}{*}{\makecell[c]{\textsc{Prompting}\\\textsc{-based}}}
    % & GPT 5 & 34.0 & \red{2.0} & 12.4 & 396k & \bluer{282k} & 7927 & \bluer{5650} & 641 & \bluer{459} \\
    % & Gemini 3 Pro & 36.0 & \red{0.0} & 12.6 & 239k & \bluer{439k} & 4779 & \bluer{8798} & 379 & \bluer{721} \\
    \midrule
    % \rowcolor{gray!10}
    & \textsc{SFT} & 36.0 & \red{6.0} & 12.3 & 678k & \redr{545k} & 13577 & \redr{10924} & 1100 & \redr{869} \\
    \rowcolor{gray!10}
    \multirow{-2}{*}{{\ares}} 
    & \textbf{\textsc{RL}} & 42.0 & -- & 11.5 & 133k & -- & 2653 & -- & 231 & -- \\
    \bottomrule
  \end{tabular}
  }
  
  \label{tab:rl}
\end{table*}

\paragraph{Training.}
To collect initial training trajectories, we utilize both publicly released agent trajectory data and trajectories collected according to the method described in Section~\ref{sec:method}. Specifically, we use \texttt{APIGen-MT}~\citep{prabhakar2025apigen} for tool-use agent training, as it provides conversational trajectories with ground truth action at every step. For \texttt{BrowseComp-Plus}, we collect trajectories using rejection sampling within our pipeline. The \texttt{WebArena} training trajectories are obtained from successful trajectories released by \textsc{AgentOccam}. For both \texttt{BrowseComp-Plus} and \texttt{WebArena}, we adopt the train-test splits following prior work~\citep{sun2025scaling,qi2024webrl,yang2024agentoccam}.

For robustness filtering in reasoning effort annotation, we test each reasoning effort level three times and use \texttt{GPT-4o} as the LLM judge, retaining only the reasoning effort that generates the correct action in all three trials. For rationale generation, we employ \texttt{GPT-5} as the rationale generator. If no reasoning effort level yields correct results across all three trials, we retain the lowest reasoning effort level with the highest accuracy. 

Regarding SFT hyperparameters, we train the {\ares} router for three epochs with a learning rate of 5e-6, a global batch size of 64, a warmup ratio of 0.1, and 0.01 weight decay. For RL training, the base model is initialized from the SFT fine-tuned model. For each prompt, we generate $G=16$ outputs to compute the group-wise advantage. We set the maximum prompt and response lengths to 4,096 and 512 tokens, respectively. During training, we use the Adam optimizer with a constant learning rate of $1.5 \times 10^{-6}$. The KL divergence coefficient is set to 0.01 to prevent the policy from deviating too far from the reference model. The total training process spans 5 epochs with a global batch size of 32.

\paragraph{Evaluation.}
For task performance, we report the average reward for \texttt{TAU-Bench}, accuracy for \texttt{BrowseComp-Plus}, and task success rate for \texttt{WebArena}. Accuracy in \texttt{BrowseComp-Plus} is evaluated using \texttt{GPT-4o} as a judge. To measure reasoning costs, we report three metrics: the total reasoning tokens generated for the entire evaluation ($T_{\text{total}}$), the average reasoning tokens generated per task ($T_{\text{task}}$), and the average tokens per inference step ($T_{\text{step}}$).

\paragraph{Baselines.}
We adopt three types of baselines: \ding{182} fixed-effort policies that consistently adopt either low or high reasoning effort, approximating the lower and upper bounds of task performance, \ding{183} a random policy that uniformly samples from the three reasoning effort levels at each step, and \ding{184} a prompting-based strategy using a large-scale reasoning LLM (\emph{e.g.}, \texttt{GPT-5}) to analyze the current state and select the appropriate reasoning effort. While using these proprietary LLMs introduces significantly higher inference costs on the router side compared to {\ares}, it provides insight into the effectiveness of prompting-based effort selection.

\begin{figure*}[t]
    \centering
    \includegraphics[width=\linewidth]{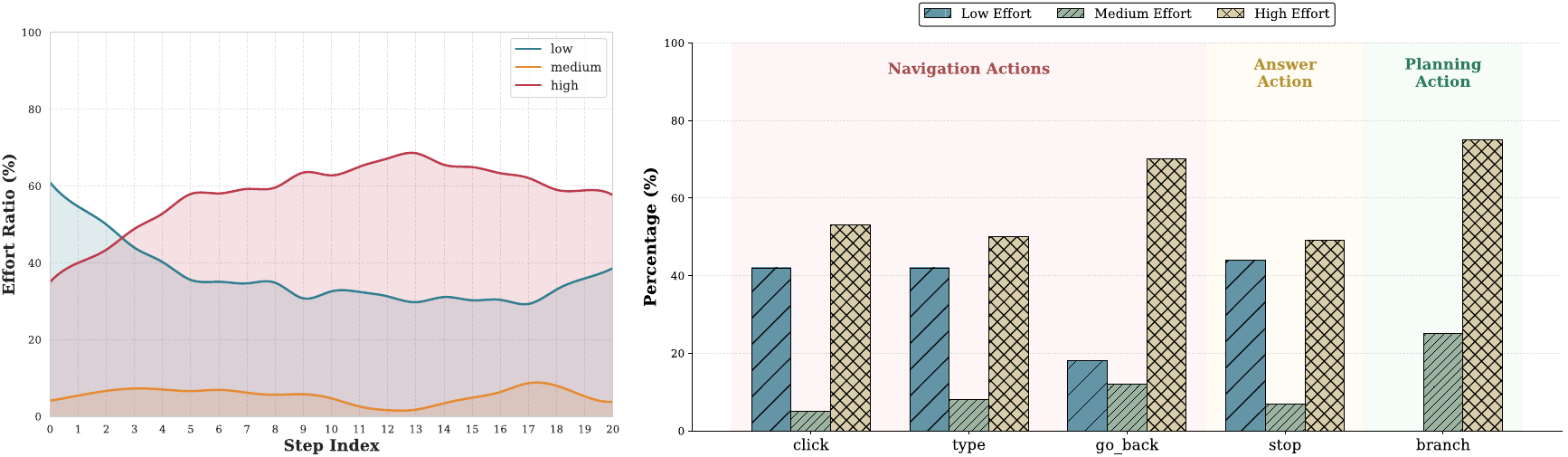} 
    
    \caption{Selection of reasoning effort by {\ares} on the \texttt{WebArena} benchmark. 
    \textbf{Left:} Percentage distribution of low, medium, and high effort levels across task step indices. 
    \textbf{Right:} Distribution of effort levels categorized by specific action types.}
    \label{fig:analysis}
\end{figure*}

\subsection{SFT Results}
In this section, we report the performance and computational efficiency of {\ares} router after SFT, across diverse agent environments.
Our main results, summarized in Table~\ref{tab:main}, demonstrate that {\ares} consistently achieves performance on par with or superior to the fixed high effort baseline while substantially reducing computational overhead. Across all benchmarks, {\ares} maintains high task success rates while delivering significant reasoning token reductions: approximately \textbf{35.2\%} on \texttt{TAU-Bench} (Retail), \textbf{41.8\%} on \texttt{BrowseComp-Plus}, and \textbf{45.3\%} on \texttt{WebArena} in total reasoning token consumption ($T_{\text{total}}$). These findings underscore that {\ares} can achieve "high-effort" results at a fraction of the cost, effectively bridging the gap between resource constraints and the performance requirements of complex agentic tasks.

We provide further analysis for each specific domain below:
\paragraph{Tool Use.} 
The evaluation on \texttt{TAU-Bench} demonstrates {\ares}'s robust domain adaptation. In the \textit{Retail} domain, {\ares} achieves a success rate of \textbf{54.8\%}, matching the High effort baseline and significantly outperforming the GPT-5 router by 14.8\%. 
In the more constrained \textit{Airline} domain, we observe a unique non-monotonic relationship where the medium strategy surpasses the high strategy, signaling an ``overthinking'' risk where excessive reasoning leads to logic drift. While the absolute performance is constrained by the backbone's capability ceiling, {\ares} effectively navigates this landscape. While {\ares} achieves parity with general-purpose routers such as \texttt{GPT-5} and \texttt{Gemini-3-Pro}. Crucially, it incurs significantly lower routing overhead compared to these larger proprietary models.

\paragraph{Deep Research.} 
On the \texttt{BrowseComp-Plus} benchmark, we observe a significant performance gap between the low (8.00\%) and high (42.7\%) effort baselines compared to other tasks. This disparity underscores that deep-research tasks are very sensitive to the reasoning effort applied. In such long-horizon tasks, a single suboptimal selection, such as choosing low effort for a complex retrieval step, can result in poorly formulated search queries. This often leads to a failure in capturing critical information, which subsequently triggers detrimental error propagation throughout the remainder of the task. Consequently, maintaining high performance requires the router to be extremely precise in identifying specific steps where reduced reasoning effort is feasible without compromising accuracy. Despite these challenges, {\ares} achieves a success rate of \textbf{41.3\%}, nearly matching the high effort ceiling while successfully reducing token consumption by \textbf{41.8\%}. This demonstrates that {\ares} effectively identifies critical reasoning nodes, ensuring task reliability while maximizing computational efficiency.

\paragraph{Web Navigation.} 
Notably, {\ares} surpasses the high effort baseline (46.5\% vs. 45.0\%) on \texttt{WebArena}, suggesting that increased reasoning effort does not consistently yield better outcomes in web navigation. We observe that excessive reasoning can lead to "overthinking," where the agent's process becomes overly divergent, resulting in practical failures such as incorrect action formats or loss of task focus. By dynamically modulating reasoning depth, {\ares} effectively mitigates these pitfalls. While {\ares} utilizes more tokens than the static configurations of \texttt{GPT-5} or \texttt{Gemini-3-Pro}, {\ares} only introduces very trivial router cost compared to them, and the significant gain in success rate highlights its superior ability to handle the dynamic nature of web environments.

\subsection{RL Results}

To evaluate the impact of our reinforcement learning phase, we compare the performance and efficiency of \textsc{Ares} trained solely with SFT against \textsc{Ares} further optimized with RL. We conduct this evaluation on both the \texttt{TAU-Bench} Retail and Airline domains. 

As shown in Table~\ref{tab:rl}, the RL optimization consistently improves both task accuracy and token efficiency. In the \emph{Retail} domain, RL training increases the success rate from 54.8\% to 58.5\% while simultaneously reducing total token consumption by 176k compared to the SFT baseline. This dual improvement is even more pronounced in the \emph{Airline} domain, where the RL-optimized router achieves a 6.0\% absolute increase in accuracy (from 36.0\% to 42.0\%) and drastically cuts total token usage by nearly 80\% (from 678k to 133k).

Furthermore, compared to the static high effort rule-based strategy, \textsc{Ares} (RL) achieves superior accuracy across both domains while consuming less than half the total reasoning tokens (476k vs. 1007k in Retail; 133k vs. 873k in Airline). These findings confirm that the RL phase effectively pushes the Pareto frontier, enabling the router to discover optimal, context-aware reasoning effort selections that the greedy SFT objective fails to capture.

\begin{figure}[t]
    \centering
    \includegraphics[width=0.8\linewidth]{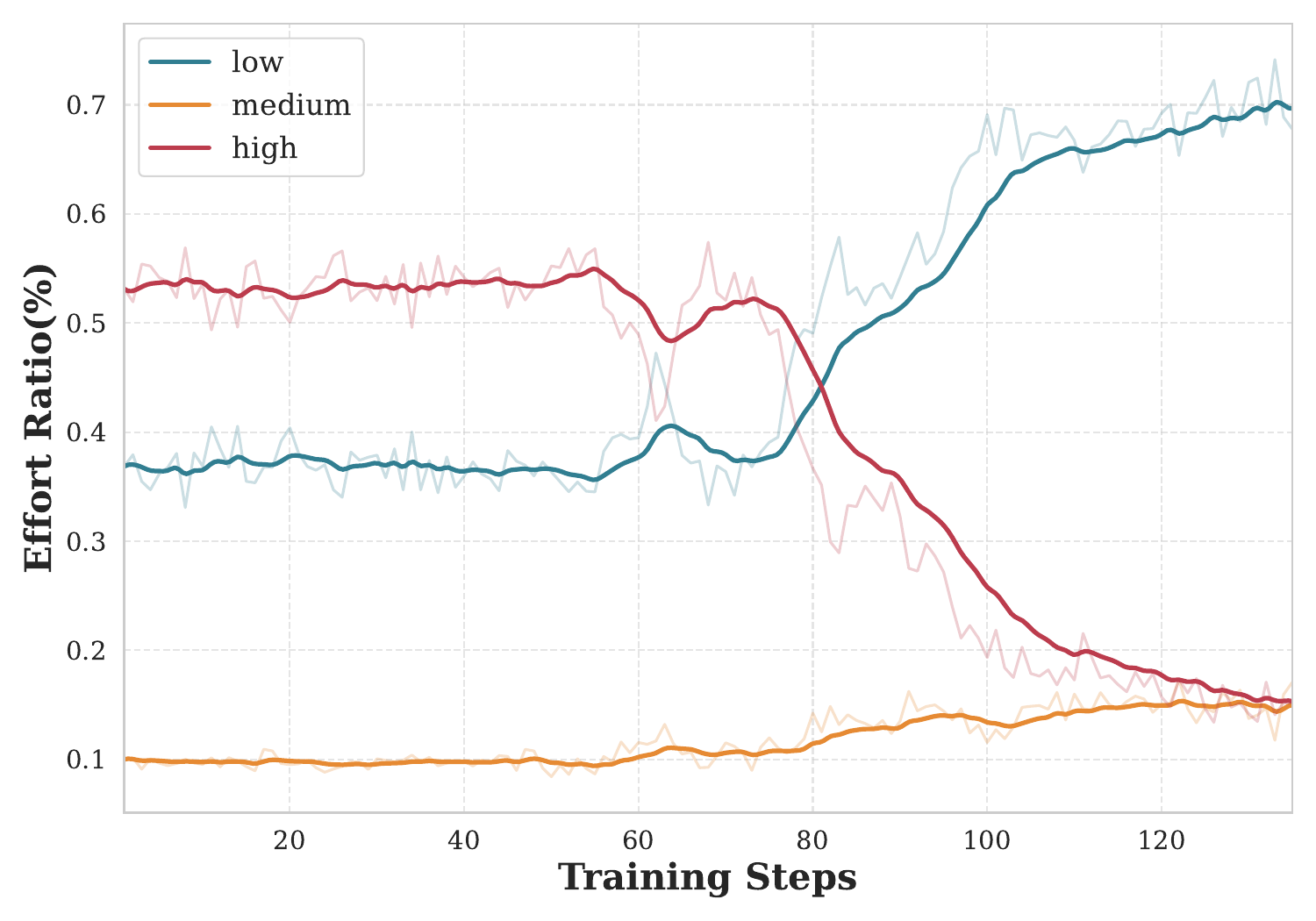} 
    \caption{Evolution of {\ares} reasoning effort prediction during GRPO training (TAU-Bench Airline).}
    \label{fig:airline_ratio}
\end{figure}

To further understand the RL optimization, we analyze the training dynamics in the \texttt{TAU-Bench} Airline domain, where the agent exhibits a counter-intuitive ``overthinking'' phenomenon: the high reasoning effort yields a lower accuracy (38.0\%) than the medium setting (42.0\%). Figure~\ref{fig:airline_ratio} illustrates how GRPO training effectively corrects this bias. While the SFT-initialized router initially selects high effort for over 50\% of the steps, the RL process rapidly suppresses this detrimental over-deliberation, dropping its usage to under 20\%. Simultaneously, the router learns to confidently rely on more efficient strategies, with the low effort ratio climbing to approximately 70\%. By autonomously curbing unnecessary computation, the RL-optimized \textsc{Ares} avoids the overthinking trap and maximizes both task success and token efficiency.

\subsection{Analysis of Reasoning Effort Selection}

To further investigate the decision-making logic of the router, we analyze the distribution of reasoning effort across two dimensions: temporal progression and action categories. Figure~\ref{fig:analysis} (left) illustrates the effort selection ratio relative to the task step index on \texttt{WebArena}. In the early stages of a task (\emph{e.g.}, steps 0--2), the router predominantly selects low reasoning effort. Our observations indicate that this is because initial navigation steps often involve low-complexity environments, such as accessing a specific forum or landing page, where the mapping from observation to action is relatively straightforward. However, as the task progresses, the frequency of high reasoning effort increases significantly.
This shift is driven by the escalating perceptual complexity of the web observations and the expanding context window, both of which impose higher cognitive demands on the agent to maintain task coherence and accuracy.

Furthermore, we evaluate how reasoning effort correlates with different action types, as shown in Figure~\ref{fig:analysis} (right). The results reveal that critical decision steps—specifically \texttt{go\_back} and \texttt{branch}—require the highest proportion of high reasoning effort. In our framework, a \texttt{go\_back} action typically signifies a strategic recovery from an incorrect navigation path, while a \texttt{branch} action indicates a substantive modification of the existing navigation plan. The high concentration of intensive reasoning at these actions suggests that {\ares} identifies these actions as high-stakes error correction or plan refinement steps. This capability demonstrates that high reasoning effort is essential for sophisticated "self-correction" mechanisms, allowing the agent to effectively navigate out of suboptimal states and re-align with the global task objective.

\subsection{Ablation Studies}
\label{subsec:ablation}
We conduct an ablation study to analyze the effects of generated rationale, SFT, and RL reward design in {\ares}. 
% The results, summarized in Table~\ref{tab:ablation}, highlight the contribution of each component to the final performance.

\begin{table}[h!]
  \centering
  \renewcommand{\arraystretch}{0.85}
  \small
  \caption{Ablation results on \textsc{Ares} components. Performance is reported on TAU-Bench Retail.}
    \resizebox{\linewidth}{!}{
  \begin{tabular}{l|cc|ccc}
    \toprule
    \rowcolor{CadetBlue!15}
    \textbf{Setting} & \textbf{Acc. (\%)} & $\Delta_{\mathrm{Acc}}$ & $T_{\text{total}}$ & $T_{\text{task}}$ & $T_{\text{step}}$ \\
    \midrule
    Low & 35.0 & \red{19.8}  & 25k & 223 & 15\\
    Medium & 47.3 & \red{7.5} & 137k & 1198 & 82 \\
    High & 54.8 & \blue{0.0} & 1007k & 8756 & 634 \\
    \midrule
    -- SFT & 41.7 & \red{13.1}  & 128k & 1113 & 74\\
    -- Rationale & 51.3 & \red{3.5} & 474k & 4128 & 285 \\
    \midrule
    {\ares}(SFT) & 54.8 & - & 652k& 5677 & 388 \\
    \bottomrule
  \end{tabular}
  }
  \label{tab:ablation}
\end{table}

\paragraph{Effect of Supervised Fine-tuning.} 
As shown in the ``- SFT'' setting in Table~\ref{tab:ablation}, removing fine-tuning and using \texttt{Qwen3-1.7B} out-of-the-box leads to the most significant degradation, with accuracy dropping from 54.8\% to 41.7\%. Although this setting has the lowest token consumption, the low accuracy indicates that without task-specific tuning, the base model lacks the judgment to allocate sufficient reasoning effort, often defaulting to overly simplistic strategies.

\paragraph{Effect of Reasoning Rationale.} 
We evaluate Phase 3 by training a router that predicts effort labels directly without rationales (``- Rationale''), resulting in a 3.5\% accuracy drop. This confirms that, by forcing the router to explicitly analyze task difficulty before deciding, the internal thinking process acts as a necessary cognitive bridge that significantly improves the accuracy of effort selection.

\paragraph{Reward Design.}
We further conduct an ablation study to analyze our RL reward design. Since a key component of the {\ares} training objective is the reasoning cost penalty, $R_{\text{cost}}$, we evaluate a variant that uses an accumulated step-wise cost without normalization, defined as $R_{\text{cost}} = \sum_{t=1}^{T} c(e_t)$. To ensure that successful trajectories still yield a net positive total reward, we proportionally scale down the step-wise penalties in this unnormalized setting to $c(e_t) = -0.02, -0.06,$ and $-0.12$ for low, medium, and high reasoning efforts, respectively.

\begin{figure}[t]
    \centering
    \includegraphics[width=\linewidth]{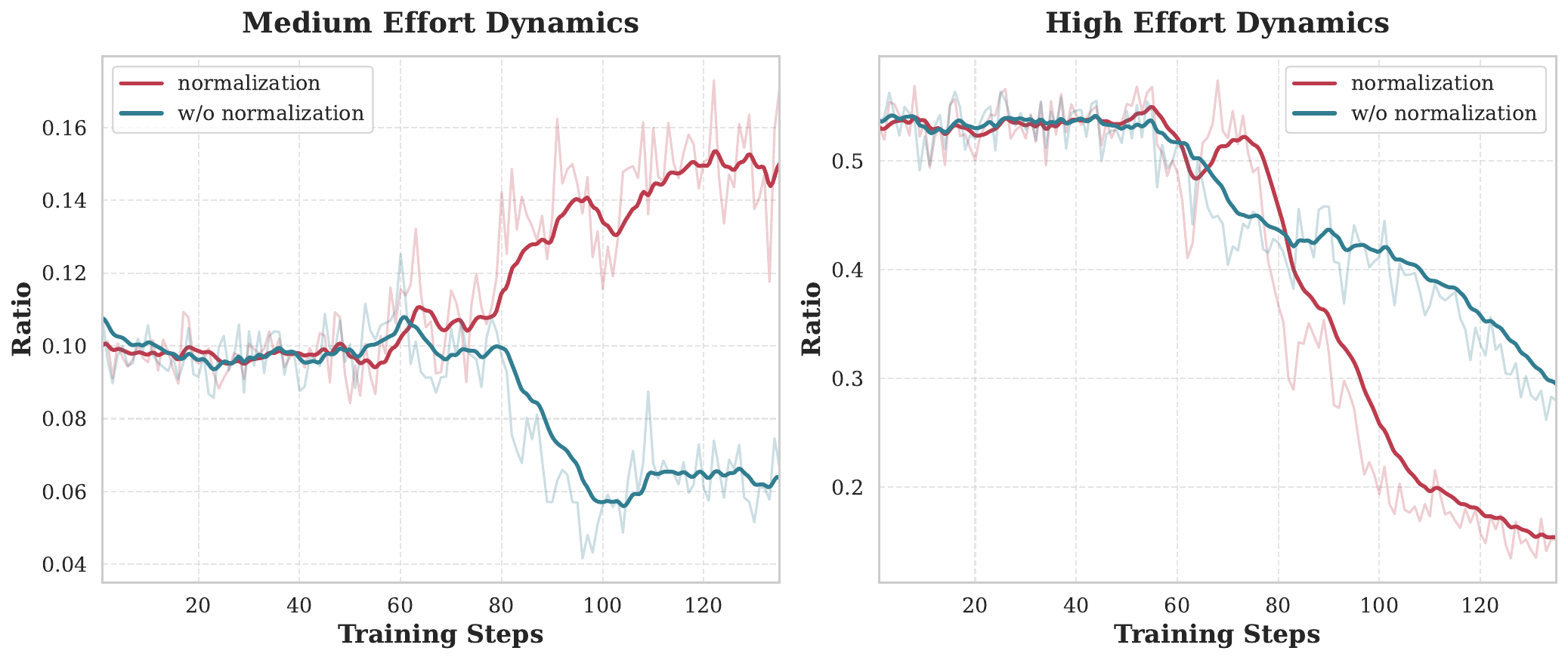} 
    
    \caption{
        Comparison of medium-effort (left) and high-effort (right) ratios between whether to use normalized reasoning cost reward during training. 
    }
    \label{fig:ablation_ratio}
\end{figure}

Figure~\ref{fig:ablation_ratio} illustrates the training dynamics on the \texttt{TAU-Bench} Airline domain. As shown in the right panel, applying normalization to the cost reward leads to a much more aggressive and effective compression of high effort usage, driving its selection ratio down to approximately 15\%, compared to 30\% in the unnormalized setting. Furthermore, since medium effort is the optimal reasoning strategy for the Airline domain, the normalized reward correctly incentivizes the router to progressively increase its reliance on this optimal effort level, as seen in the left panel. Conversely, without normalization, the usage of medium effort declines after 80 training steps.

\begin{table}[h!]
  \centering
  \renewcommand{\arraystretch}{0.85}
  \small
  \caption{Ablation results on \textsc{Ares} components. Performance is reported on TAU-Bench Airline.}
    \resizebox{\linewidth}{!}{
  \begin{tabular}{l|cc|ccc}
    \toprule
    \rowcolor{CadetBlue!15}
    \textbf{Setting} & \textbf{Acc. (\%)} & $\Delta_{\mathrm{Acc}}$ & $T_{\text{total}}$ & $T_{\text{task}}$ & $T_{\text{step}}$ \\
    \midrule
    Low & 32.0 & \red{10.0}  & 12k & 250 & 23\\
    Medium & 42.0 & \blue{0.0} & 98k & 1961 & 148 \\
    High & 36.0 & \red{6.0} & 873k & 17472 & 1654 \\
    \midrule
    w/o normalization & 41.3 & \red{0.7}  & 157k & 3140 & 312 \\
    normalization & 42.0 & -- & 133k & 2653 & 231 \\
    \bottomrule
  \end{tabular}
  }
  \label{tab:ablation_reward}
\end{table}

These training dynamics are directly reflected in the final evaluation results presented in Table~\ref{tab:ablation_reward}. By utilizing the normalized reward, the router not only achieves a higher task accuracy but also further reduces the reasoning cost, consuming approximately 15\% fewer total tokens (133k vs. 157k).
This indicates that an unnormalized, accumulated penalty fails to properly balance the step-wise costs against the final outcome, causing the router to struggle in converging toward the most efficient and effective effort allocation.

\subsection{Generalization Evaluation}
To assess cross-scale generalization, we evaluate {\ares} on \texttt{TAU-Bench} Retail using a \texttt{gpt-oss-120b} backbone. Since the router was trained solely on the significantly smaller \texttt{gpt-oss-20b}, this experiment tests the framework's robustness to shifts in the underlying agent's capabilities.

\begin{table}[h!]
  \centering
  \renewcommand{\arraystretch}{0.8}
  \footnotesize
  \caption{Results in TAU-Bench Retail, using \texttt{gpt-oss-120b} as the backbone LLM.}
  \begin{tabular}{l|cc|ccc}
    \toprule
    \rowcolor{CadetBlue!15}
    \textbf{Setting} & \textbf{Acc. (\%)} & $\Delta_{\mathrm{Acc}}$ & $T_{\text{total}}$ & $T_{\text{task}}$ & $T_{\text{step}}$ \\
    \midrule
    Low & 49.4 & \red{15.8}  & 26k & 231 & 17\\
    Medium & 62.0 & \red{3.2} & 111k & 972 & 74 \\
    High & 67.8 & \blue{2.6} & 558k & 4856 & 380 \\
    \midrule
    {\ares} & 65.2 & - & 428k & 3729 & 293 \\
    \bottomrule
  \end{tabular}
  \label{tab:generalize}
\end{table}

As shown in Table~\ref{tab:generalize}, {\ares} demonstrates strong cross-scale generalization. Despite the backbone being six times larger than the training source, the router achieves 65.2\% accuracy, significantly outperforming Low and Medium effort baselines. Notably, {\ares} recovers the majority of the High effort performance (67.8\%) while reducing token consumption by approximately 23\%. These results suggest that the reasoning rationales and effort-selection cues learned by {\ares} are scale-invariant, enabling effective efficiency-accuracy trade-offs even for much more capable agents.

\section{Conclusion}

We introduced {\ares}, a framework that optimizes LLM agent efficiency by dynamically selecting appropriate reasoning effort levels based on task complexity. Experimental results on benchmarks like \texttt{BrowseComp-Plus} and \texttt{TAU-Bench} demonstrate that {\ares} achieves performance parity with high-effort strategies while significantly reducing reasoning token consumption. Furthermore, our evaluation confirms that the learned reasoning patterns generalize effectively across different model scales and data sources. Future work will extend {\ares} to more diverse deployment settings, such as multi-modal inputs, to further improve inference efficiency.

% \newpage
\section*{Acknowledgement}
The work of Jingbo Yang, Bairu Hou and Shiyu Chang was partially supported by National Science Foundation (NSF) Grant IIS-2338252, and NSF Grant IIS-2302730.

\section*{Impact Statement}

This paper presents {\ares}, a framework designed to advance the field of Machine Learning by optimizing the efficiency and performance of large language model agents. By dynamically selecting appropriate reasoning effort levels, our work contributes to reducing the overall computational cost and energy consumption associated with deploying complex AI agents in real-world environments. While there are many potential societal consequences of advancing autonomous agents, we believe these implications are consistent with broader developments in the field and do not require specific highlighting here beyond standard ethical considerations.

% In the unusual situation where you want a paper to appear in the
% references without citing it in the main text, use \nocite
% \nocite{langley00}

\bibliography{example_paper}
\bibliographystyle{icml2026}

%%%%%%%%%%%%%%%%%%%%%%%%%%%%%%%%%%%%%%%%%%%%%%%%%%%%%%%%%%%%%%%%%%%%%%%%%%%%%%%
%%%%%%%%%%%%%%%%%%%%%%%%%%%%%%%%%%%%%%%%%%%%%%%%%%%%%%%%%%%%%%%%%%%%%%%%%%%%%%%
% APPENDIX
%%%%%%%%%%%%%%%%%%%%%%%%%%%%%%%%%%%%%%%%%%%%%%%%%%%%%%%%%%%%%%%%%%%%%%%%%%%%%%%
%%%%%%%%%%%%%%%%%%%%%%%%%%%%%%%%%%%%%%%%%%%%%%%%%%%%%%%%%%%%%%%%%%%%%%%%%%%%%%%
\newpage
\appendix
\onecolumn
\section{Prompts.}
In this section, we display the prompts used for the router, reasoning effort annotator, and rationale generator in {\ares}. We include the prompt used in \texttt{WebArena} as an example.

\begin{tcolorbox}[enhanced jigsaw, breakable, title=Router, colback=gray!3, colframe=blue!50]
\ttfamily
You are a router that selects the reasoning\_effort (low, medium, or high) for the OSS model's next response.

Context: The input is the current query of an OSS-model-based web agent completing the objective, including the objective, web page observation, previous interaction history, etc. The OSS model needs to decide the action for the next step and eventually complete the objective. The OSS model incurs higher cost and latency at higher reasoning levels.

- Choose HIGH if the next step likely needs careful reasoning, or complex tool sequencing.

- Choose MEDIUM if the next step is moderately complex.

- Choose LOW if the next step is straightforward and the risk is low.

Produce exactly one word: low, medium, or high.
\end{tcolorbox}

\begin{tcolorbox}[enhanced jigsaw, breakable, title=Reasoning Effort Annotator, colback=gray!3, colframe=blue!50]
\ttfamily
You are a strict evaluator of web navigation actions.

Two actions match if and only if they are exactly identical after
trimming leading and trailing whitespace and collapsing consecutive
whitespace characters into a single space.

You must enforce exact equivalence. Do NOT allow any differences in:

- action type or name,

- arguments or argument values,

- URLs, selectors, or identifiers.

If any difference exists, the actions must be judged as NOT matching.

\end{tcolorbox}

\begin{tcolorbox}[enhanced jigsaw, breakable, title=Rationale Generator, colback=gray!3, colframe=blue!50]
\ttfamily
You are annotating router training data for a web-navigation agent.

Background: The router selects the reasoning\_effort (low, medium, high) for an OSS model's next action. Higher effort means more careful reasoning and higher cost/latency. The context includes the user's objective and the agent's interaction history leading up to the next action.

Task: Given the context, the chosen label, as well as OSS model response from different efforts, write up to five concise sentences
explaining why this label is appropriate. Ground your explanation in the
objective and interaction history in the given context. Reveal chain-of-thought
or step-by-step reasoning for why you choose the corresponding reasoning effort through analyzing current task states and predicting the difficulty of next sub-task.

\end{tcolorbox}
%%%%%%%%%%%%%%%%%%%%%%%%%%%%%%%%%%%%%%%%%%%%%%%%%%%%%%%%%%%%%%%%%%%%%%%%%%%%%%%
%%%%%%%%%%%%%%%%%%%%%%%%%%%%%%%%%%%%%%%%%%%%%%%%%%%%%%%%%%%%%%%%%%%%%%%%%%%%%%%

\section{Training Example.}
In this section, we provide a training example used for SFT.

\begin{tcolorbox}[enhanced jigsaw, breakable, title=Training Example, colback=gray!3, colframe=red!50]
\ttfamily
\textbf{System Prompt:}
You are a router that selects the reasoning\_effort (low, medium, or high) for the OSS model's next action.

Context: One web agent interacting with web elements to complete objectives. The OSS model charges more cost and latency at higher reasoning levels.

- Choose HIGH if the next step likely needs careful reasoning, or the current observation contains complex patterns.

- Choose MEDIUM if the next step is moderately complex.

- Choose LOW if the next step is straightforward and the risk is low.

Produce exactly one word: low, medium, or high.

\textbf{User Prompt:}
OBJECTIVE:
Where is the nearest In-N-Out to Upitts, and what is the walking distance to it?

INTERACTION HISTORY:
Step 0:
REASON: To proceed with finding the nearest In-N-Out to Upitts, the first logical step is to use the search function provided by OpenStreetMap. By entering the query for the nearest In-N-Out to Upitts in the search textbox and submitting it, we can potentially gather information on the nearest location. This will help in identifying the location before proceeding to calculate the walking distance.
ACTION: type [35] [In-N-Out near University of Pittsburgh] [1]
OBSERVATION: RootWebArea [1] `OpenStreetMap'
heading `OpenStreetMap logo OpenStreetMap'
link [41] `OpenStreetMap logo OpenStreetMap'
link [402] `Edit'
button [403]
link [373] `History'
link [374] `Export'
link [407] `GPS Traces'
link [408] `User Diaries'
link [409] `Communities'
link [410] `Copyright'
link [411] `Help'
link [412] `About'
link [382] `Log In'
link [383] `Sign Up'
link [515] `Where is this?'
textbox [35] `Search'
button [516] `Go'
link [503] `Find directions between two points'
heading `Welcome to OpenStreetMap!'
button [469] `Close'
text `OpenStreetMap is a map of the world, created by people like you and free to use under an open license.'

\textbf{Label:} High

\textbf{Rationale:}
The objective requires finding the nearest In-N-Out relative to the University of Pittsburgh and computing a walking distance, which likely involves interpreting search results, verifying the correct location, and using the directions tool. The current page exposes multiple relevant actions, such as search and route planning, that require careful selection. Since In-N-Out locations are uncommon near Pittsburgh, identifying the correct nearest location is nontrivial and demands careful reasoning. Therefore, a high reasoning effort is appropriate.

\end{tcolorbox}

\section{Dataset Statistics}
In the Table~\ref{tab:dataset}, we provide the training data statistics.

\begin{table}[t]
  \centering
  \small
  \setlength{\tabcolsep}{6pt}
  \renewcommand{\arraystretch}{1.15}
  \caption{Dataset statistics of reasoning-effort labels across benchmarks.}
  % \resizebox{0.5\linewidth}{!}{
  \begin{tabular}{l|r|rrr}
    \toprule
    \textbf{Dataset} & \textbf{Total} & \textbf{High} & \textbf{Medium} & \textbf{Low} \\
    \midrule
    TAU-Bench        & 43{,}358 & 12{,}261 & 6{,}222 & 24{,}875 \\
    BrowseComp-Plus  & 12{,}366 & 6{,}184  & 3{,}091 & 3{,}091  \\
    WebArena         & 1{,}718  & 1{,}095  & 72      & 551      \\
    \bottomrule
  \end{tabular}
  \label{tab:dataset}
\end{table}

\end{document}